\pdfoutput=1
\documentclass[11pt]{article}

\usepackage[]{emnlp2021}

\usepackage{times}
\usepackage{latexsym}
\usepackage{graphicx}
\usepackage{tabularx}
\usepackage{amsfonts}
\usepackage[]{url}

\usepackage{fancyhdr}

\usepackage[T1]{fontenc}
\usepackage[utf8]{inputenc}

\title{Adapting Language Models for Zero-shot Learning by Meta-tuning on Dataset and Prompt Collections}

\author{
Ruiqi Zhong \quad Kristy Lee$^{*}$ \quad Zheng Zhang$^{*}$ \quad Dan Klein \\
Computer Science Division, University of California, Berkeley \\
\{ruiqi-zhong, kristylee, zhengzhang1216, klein\}@berkeley.edu
}

\begin{document}
\maketitle

\begin{abstract}
%The goal of zero-shot learning is to produce answers based on task descriptions.
%For instance, to classify review sentiments, we ask the model to answer ``\textit{Yes}"/``\textit{No}" to the question ``\textit{Is the review positive?}".
%For example, we can ``prompt" the language model with the review and the question ``\textit{Is the review positive?}", and ask whether the next word is ``\textit{Yes}" or ``\textit{No}".
Large pre-trained language models (LMs) such as GPT-3 have acquired a surprising ability to perform zero-shot learning.
For example, to classify sentiment without any training examples, we can ``prompt" the LM with the review and the label description ``\textit{Does the user like this movie?}", and ask whether the next word is ``\textit{Yes}" or ``\textit{No}".
However, the next word prediction training objective is still \textbf{misaligned} with the target zero-shot learning objective. 
To address this weakness, we propose meta-tuning, which directly optimizes the zero-shot learning objective by fine-tuning pre-trained language models on a collection of datasets.
We focus on classification tasks, and construct the meta-dataset by aggregating 43 existing datasets and annotating 441 label descriptions in a question-answering (QA) format.
When evaluated on unseen tasks, meta-tuned models outperform a same-sized QA model and the previous SOTA zero-shot learning system based on natural language inference.
Additionally, increasing parameter count from 220M to 770M improves AUC-ROC scores by 6.3\%, and we forecast that even larger models would perform better. 
Therefore, measuring zero-shot learning performance on language models out-of-the-box might underestimate their true potential, and community-wide efforts on aggregating datasets and unifying their formats can help build models that answer prompts better.

\end{abstract}

\section{Introduction}

The goal of zero-shot classification (ZSC) is to classify textual inputs using label descriptions without any examples \cite{yin-etal-2019-benchmarking}. 
Large language models - whose only training objective is to predict the next word given the context - have acquired a surprising ability to perform ZSC \cite{radford2019language, brown2020language, le-scao-rush-2021-many}.
For example, to classify whether the sentence ``\textit{This movie is amazing!}" is positive, we can prompt the language model with the context ``\textit{Review: This movie is amazing! Positive Review? \_\_\_ }", and check whether the next word is more likely to be ``\textit{Yes}" or ``\textit{No}" \cite{Zhao2021Calibrate}.
To convert ZSC into a language modeling (LM) task that an LM model is likely to perform well, many recent works focus on finding better prompts \cite{shin-etal-2020-autoprompt, schick2020exploiting, schick2020s, gao-etal-2021-making}.

However, the LM training objective is correlated but still misaligned with the target objective to answer prompts.
\iffalse
For example, to classify review sentiment, we can ask the model to answer ``\textit{Yes/No}" to the question ``\textit{Does the user like this movie}?" (Figure \ref{fig:fig1} (a)).
UnifiedQA \cite{khashabi-etal-2020-unifiedqa}, a model trained to answer generic questions and initialized with T5 (770 million parameters) \cite{raffel2019exploring}, can achieve zero-shot accuracy 92\% on SST-2 \cite{socher2013recursive} by answering this question, while GPT3 (175 B parameters) only obtains 80\% accuracy \cite{Zhao2021Calibrate}.
Being adapted to QA makes a 200x smaller model answer prompts better. 
\fi
Our work addresses this weakness by directly optimizing the zero-shot classification objective through fine-tuning (Section \ref{sec:training}).
This requires us to 1) unify different classification tasks into the same format, and 2) gather a collection of classification datasets and label descriptions (prompts) for training (Section \ref{main-sec:data}).
Since we fine-tune our model on a meta-dataset, we name our approach meta-tuning.
%Our method lies in between the two extremes of 1) prompting \textit{general}-purpose language models out of the box to perform ZSC, and 2) fine-tuning them on \textit{specific} tasks.
%In other words, we make the model specialized in ZSC, but still able to generalize to new unseen tasks.

\begin{figure*}[t!]
    \centering
    \includegraphics[width=0.9\textwidth]{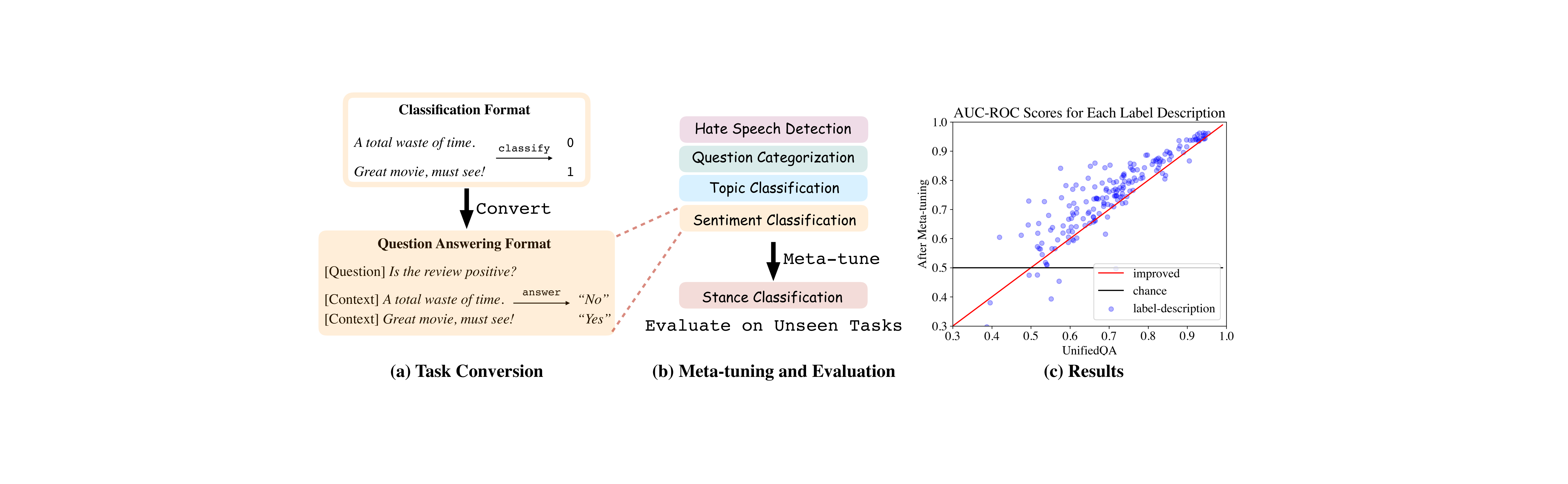}
    \caption{\textbf{(a)} We convert the format to question answering. We manually annotate label descriptions (questions) ourselves (Section \ref{main-sec:data}). \textbf{(b)} We finetune the UnifiedQA \cite{khashabi-etal-2020-unifiedqa} model (with 770 M parameters) on a diverse set of tasks (Section \ref{sec:training}), and evaluate its 0-shot classification (ZSC) performance on an unseen task. \textbf{(c)} For each label description (question) we evaluate the AUC-ROC score for the ``Yes" answer, and each dot represents a label description (Section \ref{sec:metrics}). The $x$-value is the ZSC performance of UnifiedQA; the $y$-value is the performance after meta-tuning. 
    In most cases, the $y$-value improves over the $x$-value (above the red line) and is better than random guesses (above the black line) by a robust margin (Section \ref{sec:results}).
    }
    \label{fig:fig1}
\end{figure*}

We focus on binary classification tasks and unify them into a ``\textit{Yes}"/``\textit{No}" QA format \cite{clark-etal-2019-boolq, mccann2018natural}, where the input is provided as the context and the label information is provided in the question (Figure \ref{fig:fig1} (a)). 
Using this format, we gathered a diverse set of classification datasets from 43 different sources listed on Kaggle, SemEval, HuggingFace, and other papers.
These tasks range from hate speech detection, question categorization, sentiment classification to stance classification, etc, and the genre ranges from textbooks, social media, to academic papers, etc.
In total, these datasets contain 204 unique labels, and we manually annotated 441 label descriptions (Figure \ref{fig:data-collection}).

To evaluate ZSC, we need to define what counts as a task that the model has not seen during training time.
While prior work considers different notions of ``unseen" by disallowing the same label or the same dataset to appear during training, 
our work defines ``unseen" more harshly by disallowing similar datasets.
For example, we consider AG News topic classification dataset \cite{10.5555/2969239.2969312} and the topic classification dataset from \citet{yin-etal-2019-benchmarking} to be similar, even though their sources and label spaces are different.

%We evaluate the UnifiedQA model before and after meta-tuning by calculating the AUC-ROC score for each label description. 
% Figure \ref{fig:fig1} (c) shows the results:
%for most label descriptions, our meta-tuned model is better and beats the 0.5 random baseline by a robust margin.
Meta-tuning improves ZSC over UnifiedQA for most labels (Figure \ref{fig:fig1} (c)).
Moreover, larger models are better, and hence we forecast that meta-tuning would work for even larger models.
We also find that the performance can be slightly improved by training on datasets similar to the test dataset, ensembling different label descriptions, or initializing with a QA model  (Section \ref{sec:core-results}). 
All of our findings reliably hold under different robustness checks (Section \ref{sec:robust}), and our approach outperforms the previous SOTA \citet{yin-etal-2019-benchmarking} using the same pre-training method (Section \ref{sec:comparison}).

Our results suggest two promising future directions (Section \ref{sec:implications}).
First, large language models' (e.g. GPT-3) potential for zero-shot learning, as currently measured by context-prompting, might have been broadly underestimated; meta-tuning might significantly improve their performance.
Second, community-wide efforts on aggregating and unifying datasets can scale up training and evaluation for zero-shot learning models.
On the flip side, however, the meta-tuning approach might incentivize providers of LM inference APIs to collect prompts from users, hence potentially leading to security, privacy, and fairness concerns at a greater scale (Section \ref{sec:ethics}).

\paragraph{Contributions} To summarize, we 1) curate a dataset of classification datasets with expert annotated label descriptions. 2) demonstrate a simple approach to train models to perform zero-shot learning, and 3) identify several factors that improve performance; in particular, larger pretrained models are better.
\footnote{Code and data available here: \url{https://github.com/ruiqi-zhong/Meta-tuning}.}

\section{Data} \label{main-sec:data}
We gather a wide range of classification datasets and unify them into the ``\textit{Yes}"/``\textit{No}" question answering format for binary classification.
Then we group similar datasets together to determine what counts as unseen tasks during evaluation. 

\paragraph{Gathering classification datasets}
We collect classification datasets from
Kaggle\footnote{\url{https://www.kaggle.com}}, 
Huggingface \cite{wolf-etal-2020-transformers},
SemEval\footnote{\url{https://semeval.github.io} }, 
and other papers. 
We looked through these sources and only considered English classification datasets. 
We also skipped the tasks that we felt were already better represented by other datasets in our collection. 
Then we manually examined a few examples in each remaining dataset to make sure it seemed plausibly clean.

The goals of these classification datasets include, but are not limited to sentiment classification (IMDB Reviews, \citet{maas-EtAl:2011:ACL-HLT2011}), topic classification (AG News, \citet{10.5555/2969239.2969312}), grammaticality judgement (CoLA, \citet{warstadt2018neural}), paraphrase detection (QQP\footnote{\url{https://www.kaggle.com/c/quora-question-pairs}}), definition detection (SemEval 2020 Task 6, \citet{spala-etal-2019-deft}), stance classification (SemEval 2016 Task 6, \citet{mohammad-etal-2016-semeval}),  etc. 
The genre includes academic papers, reviews, tweets, posts, messages, articles, and textbooks.
The comprehensive list of datasets is in Appendix \ref{appendix-sec:dataset}.
Overall, we aim for a high diversity of tasks and genres by building upon what the broader research community has studied. 
Our approach is complementary to that of \citet{weller-etal-2020-learning}, which asks turkers to generate tasks, and that of \citet{mishra2021natural}, which generates tasks by decomposing existing templates used to construct reading comprehension datasets.
The concurrent work of \citet{bragg2021flex} unifies the evaluation for few-shot learning;
their zero-shot evaluation setup is the closest to ours, and they used templates and verbalizers \cite{schick2020exploiting} to specify the semantics of a task. 

Some of our datasets are noisy and not peer reviewed, or contain tasks that are too complicated (e.g. Multi-NLI, \citet{williams-etal-2018-broad}) for ZSC.
To make our evaluation more informative, we only include them for training but not testing. 
We make these decisions before running our experiments in Section \ref{sec:results} to prevent selection bias. 

\begin{figure}
    \centering
    \includegraphics[width=0.44\textwidth]{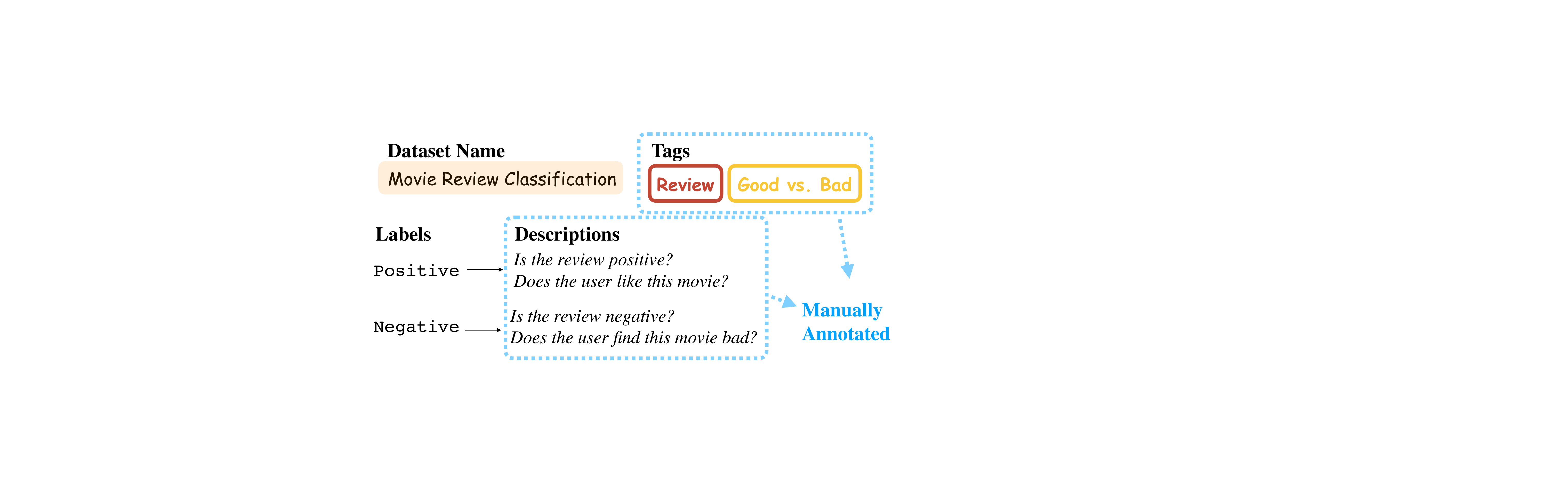}
    \caption{For each dataset, we annotate 1-3 descriptions for each label in the form of questions, and associate it with a set of property tags. The question answering format can be seen in Figure \ref{fig:fig1} (a).
    }
    \label{fig:data-collection}
\end{figure}

\paragraph{Unifying the dataset format}

\begin{figure}
    \centering
    \includegraphics[width=\columnwidth]{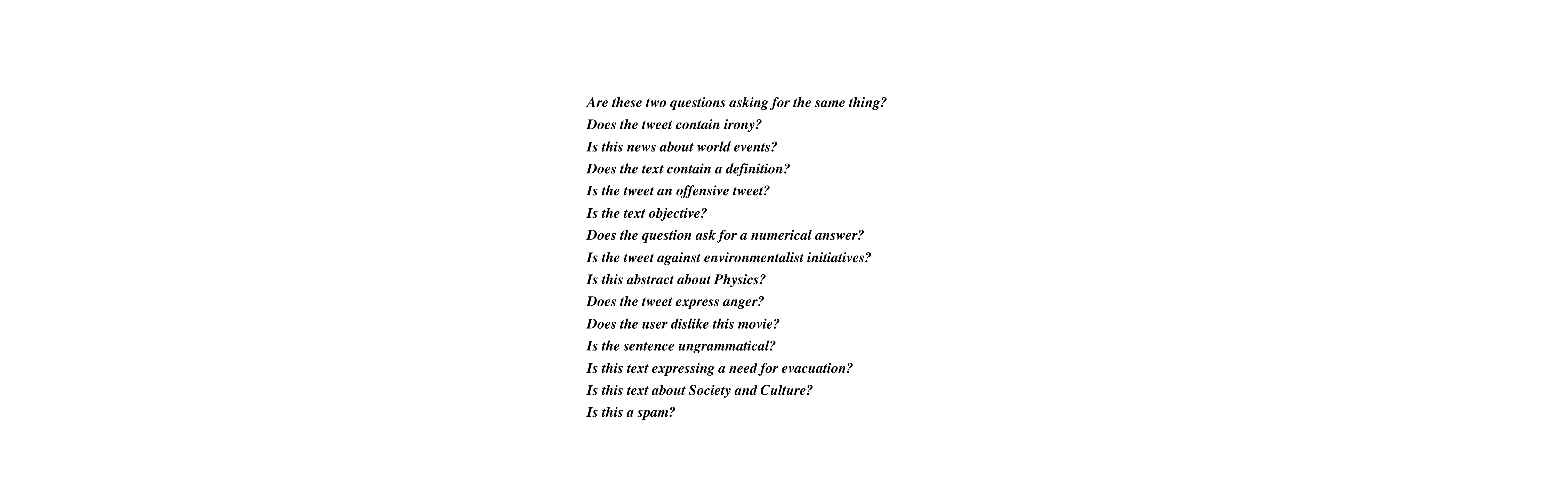}
    \caption{Some example manually annotated label descriptions (questions). Three of the authors manually wrote 441 questions in total, and each of them is proof-read by at least another author.}
    \label{fig:example-descriptions}
\end{figure}

We convert each classification dataset into a \textit{``Yes"/``No"} question answering format and provide label information in the question.
For each label, we annotate 1-3 questions. 
If the label is null (for example, a text that does not express a particular emotion in an emotion classification dataset), we skip this label. 
Three of the authors\footnote{One of them is a graduate student and the other two are undergrads; all of them study Computer Science and have taken an NLP class.} manually annotated 441 questions for 204 unique labels, and each question is proofread by at least another author. 
See Figure \ref{fig:data-collection} for a concrete example, and Figure \ref{fig:example-descriptions} for some representative label descriptions. 

Additionally, some datasets contain thousands of labels \cite{chalkidis-etal-2019-large, allaway-mckeown-2020-zero}.
In this case, we use templates to automatically synthesize label descriptions and exclude them from evaluation.

\paragraph{Grouping similar datasets}
Our goal is to test the models' ability to generalize to tasks that are different enough from the training tasks.
Therefore, at test time, we need to exclude not only the same dataset that appeared in the meta-tuning phase, but also ones that are similar.

This poses a challenge: whether two datasets perform the same task involves subjective opinion, and there is no universally agreed definition.
On one extreme, most datasets can be counted as dis-similar tasks, since they have different label spaces and input distributions.
On the other extreme, all datasets can be considered the same task, since they can all be unified into the question answering format.

\begin{figure}
    \centering
    \includegraphics[width=\columnwidth]{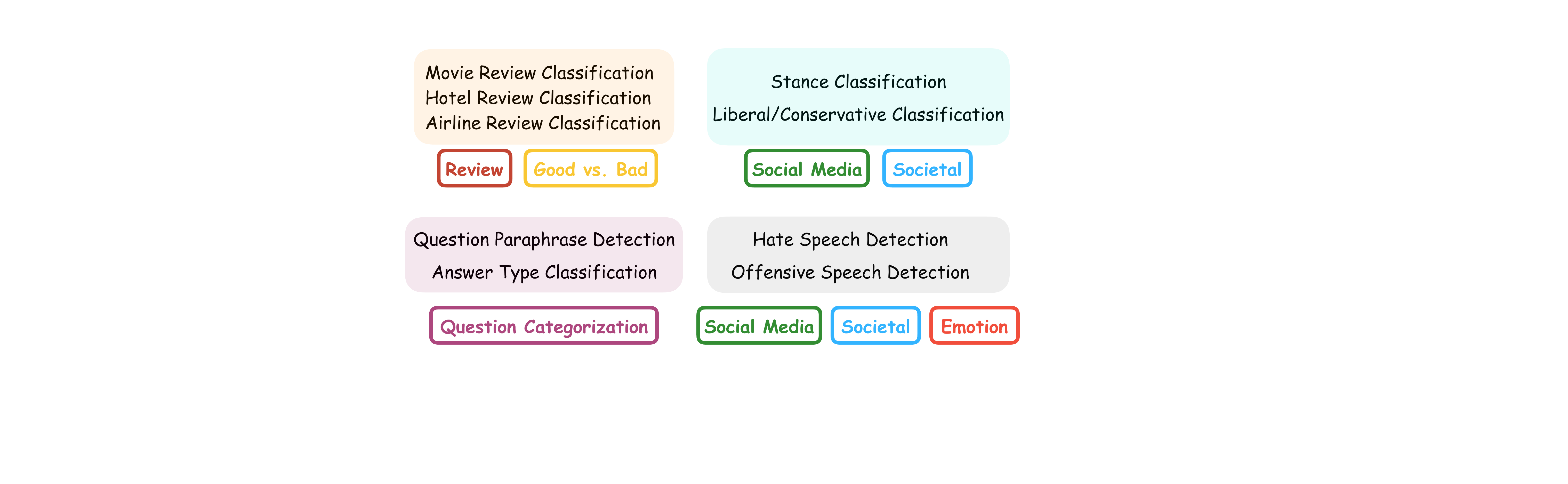}
    \caption{Example dataset groups based on tags. We never train and test on datasets from the same group, e.g. train on hotel review and test on movie review. }
    \label{fig:task-group}
\end{figure}

To tackle this challenge, we create a set of tags, each describing a dataset property. 
The set of tags includes \textit{domain classification}, \textit{article}, \textit{emotion}, \textit{social-media}, etc, and the full set of them can be seen in Appendix \ref{appendix-sec:dataset-property}.
Then we define the two datasets to be similar if they are associated with the same set of tags, and prohibit the model to learn from one and test on the other.
For example, our work considers the topic classification datasets from \citet{10.5555/2969239.2969312} (AG News) and \citet{yin-etal-2019-benchmarking} to be similar since they both classify topics for articles, even though their sources and label spaces are different.
Some example dataset groups can be seen in Figure \ref{fig:task-group}.

Nevertheless, our procedure is not bullet-proof and one can argue that our notion of unseen tasks, though harsher than prior works \cite{yin-etal-2019-benchmarking, pushp2017train}, is still lenient. 
Therefore, as additional robustness checks, for each dataset we evaluate, we manually identify and list the most relevant dataset that is allowed during training in Appendix \ref{appendix-sec:most-relevant} . 
For example, the most relevant dataset to the IMDB review sentiment classification dataset is the emotion classification dataset from \citet{yin-etal-2019-benchmarking}, which classifies the input text into 9 emotions, such as ``\textit{joy}", ``\textit{surprise}", ``\textit{guilt}", etc. 
We consider the emotion classification dataset to be relevant, since sentiment classification often involves identifying emotions. 
However, one can also argue that they are different tasks: their input and label spaces are different, and sadness can be caused by a great tragedy, or a bad movie that wastes the users' time.
The comprehensive list of label descriptions grouped by dataset similarity is in Appendix \ref{appendix-sec:label-descriptions}.

In total, we spend around 200 hours to collect this dataset.
This time estimate includes skimming through the dataset repos and recent NLP papers, writing programs to download the datasets and unify their format, annotating label descriptions, performing quality controls, and documenting the collection process.
\section{Metrics} \label{sec:metrics}
To reliably aggregate performance across different datasets and present as much information as possible, we report a set of descriptive statistics and provide visualizations whenever we compare two models.
We generally do not reduce a model's performances on different datasets into one scalar quantity and compare this number only.

\paragraph{Descriptive statistics}
For each label description (question), we calculate the AUC-ROC score
\footnote{We do not evaluate F-score or accuracy, since they are very sensitive to the decision cutoff, and usually additional calibration is needed \cite{Zhao2021Calibrate}.} 
by treating the ``\textit{Yes}" answer as the positive class. 
After calculating the AUC-ROC score for each label, we calculate the following set of descriptive statistics to compare two models. 
Suppose that model $Y$ is hypothetically better than $X$. 
Denoting $\Delta$ as the change of AUC-ROC of a label description from $X$ to $Y$, we can summarize how $\Delta$ is distributed across the set of label descriptions with the following statistics:

\begin{itemize}
    \item $\mathbb{E}[\Delta]$: the average change in AUC-ROC.
    \item $\mathbb{P}[\Delta > t]$: the fraction of label descriptions where the change is over the threshold $t$. 
    \item $\mathbb{P}[\Delta < -t]$: the fraction of label descriptions where the change is less than $-t$. 
    \item $Std[\Delta]$: the standard deviation of the change.
\end{itemize}

In the main paper, we weight each label description equally in this distribution to calculate the above statistics. 
We may also weight each label or dataset equally, and the corresponding results are in Appendix \ref{appendix-sec:comprehensive-results}.
To make sure our conclusions are robust, we consider one model to be better only when $\mathbb{E}[\Delta] > 0$ and $\mathbb{P}[\Delta > t] > \mathbb{P}[\Delta < -t]$ for all $t \in \{1\%, 5\%, 10\%\}$, under all three types of weighting.
In other words,  we claim that one model is better than the other only when 12 conditions simultaneously hold. 

\paragraph{Visualizations} We use scatter plots to visualize and compare the performance of two models, where each dot represents a label description, its x-value represents the AUC-ROC score of the model $X$, and its y-value represents that of $Y$.
If most dots are above the identity line $y=x$, the model $Y$ is better than $X$. 

\begin{figure}
    \centering
    \includegraphics[width=\columnwidth]{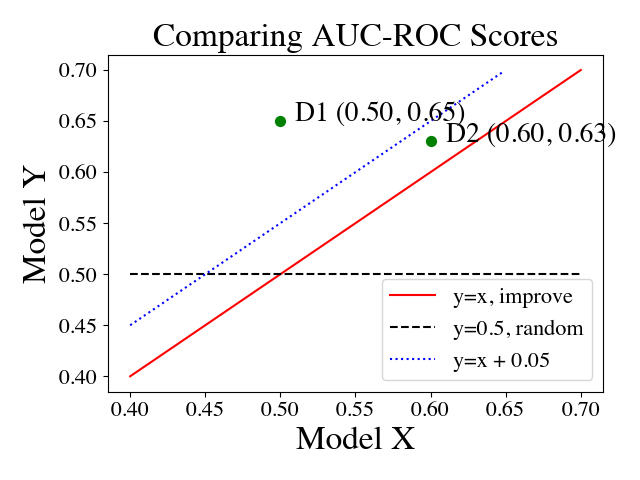}
    \caption{Each dot represents a label description, and its $x/y$-value each represents the performance of model $X/Y$ (measured by AUC-ROC score). For example, on label description $D1$, model $X/Y$ has AUC-ROC score 0.5/0.65. If the dot is above the black line ($y = 0.5$), model $Y$ is performing better than random guesses. If the dot is above the red line ($y = x$), model $Y$ is better than model $X$.
    Since one out of two dots are above $y = x + 0.05$, we have $\mathbb{P}[\Delta > 5\%] = 0.5$.}
    \label{fig:metrics-explained}
\end{figure}

The descriptive statistics and the visualizations are explained in Figure \ref{fig:metrics-explained}.

\section{Model} \label{sec:training}

\paragraph{Architecture} 
We format the inputs to the model in the same way as UnifiedQA \cite{khashabi-etal-2020-unifiedqa}, which concatenates the context to the question and adds a ``\textit{[SEP]}" token in between. 
Then we feed the concatenated input into the T5 encoder and produce the answer score by normalizing the ``\textit{Yes}"/``\textit{No}" probability of the first decoded token.
Unless otherwise noted, we initialize our model with T5-Large (770 Million parameters). 
We sometimes compare to or initialize with the UnifiedQA model \cite{khashabi-etal-2020-unifiedqa}, which is trained on a wide range of question answering datasets.
For a fair comparison, we use the UnifiedQA model initialized with T5-Large as well.
To meta-tune non-Seq2Seq pre-trained models, such as BERT \cite{devlin-etal-2019-bert} or RoBERTa \cite{liu-etal-2019-robust}, we add an MLP layer on top of the pooled output/``\textit{[CLS]}" token to classify between ``\textit{Yes}"/``\textit{No}". 
We leave the improvement on model architectures \cite{ye2021zero, li2021prefix, lester2021power} and training objectives \cite{murtydreca, Yin2020Meta-Learning} for future work. 

\paragraph{Meta-tuning}
We create a training distribution that balances between datasets, label descriptions, and ``\textit{Yes}"/``\textit{No}" answers. 
To create the next training datapoint for meta-tuning, we select a dataset from the training split uniformly at random (u.a.r.); then we select a label description (question) u.a.r. and with 50\% probability select a textual input with the answer ``\textit{Yes}"/``\textit{No}".
To prevent over-fitting, we do not train on any combination of label description and textual input twice.
Unless otherwise noted, we meta-tune the model for 5000 steps and use batch size 32.
We did not tune any hyper-parameters or training configurations since they work well during our first attempt.
To evaluate ZSC performance on each dataset, we leave out one group of similar datasets as the evaluation set and train on the rest.
Altogether, the experiments take around 250 GPU hours on Quadro 8000. 

\section{Results} \label{sec:results}

\subsection{Hypotheses and Conclusions} \label{sec:core-results}

\begin{table*}[t!]
    \centering
    \begin{tabular}{lcccc}
    \hline
{} & $\mathbb{E}[\Delta]$ & $\mathbb{P}[\Delta > 1\%]$ &  $\mathbb{P}[\Delta < -1\%]$ &  $Std(\Delta)$ \\
\hline
Meta-tuned vs. UnifiedQA    &            3.3\% &            59.5\% &             28.1\% &        9.5\% \\
Larger &            6.3\% &            75.1\% &             15.1\% &        8.1\% \\
Pre-trained vs. Random &   23.8\% & 95.7\% &              3.2\% &            14.0\% \\
Train on Similar  &            0.7\% &            43.8\% &             20.5\% &        3.2\% \\
Ensemble Descriptions  &            0.7\% &            28.9\% &             16.8\% &        3.1\% \\
Initialize with UnifiedQA            &            1.1\% &            54.1\% &             24.3\% &        6.9\% \\
    \hline
    \end{tabular}
    \caption{The statistics used to compare two models, introduced in Section \ref{sec:metrics}. The larger $\mathbb{E}[\Delta]$ and the difference between $\mathbb{P}[\Delta > 1\%]$ and $\mathbb{P}[\Delta < -1\%]$, the better. Row 1 finds that a meta-tuned model is better than UnifiedQA; row 2 finds that the larger model is better; row 3 finds that pre-training does the heavy lifting; row 4, 5, and 6 finds that the performance can be improved by training on similar datasets, ensembling label descriptions, and initializing with a UnifiedQA model. 
    Note that $Std(\Delta)$ is the standard deviation of individual descriptions, not the standard deviation of the estimated mean.
    Due to space constraint we only show $t=1\%$ in this table.}
    \label{tab:main-tab}
\end{table*}

\begin{figure*}[]
    \centering
    \includegraphics[width=0.9\textwidth]{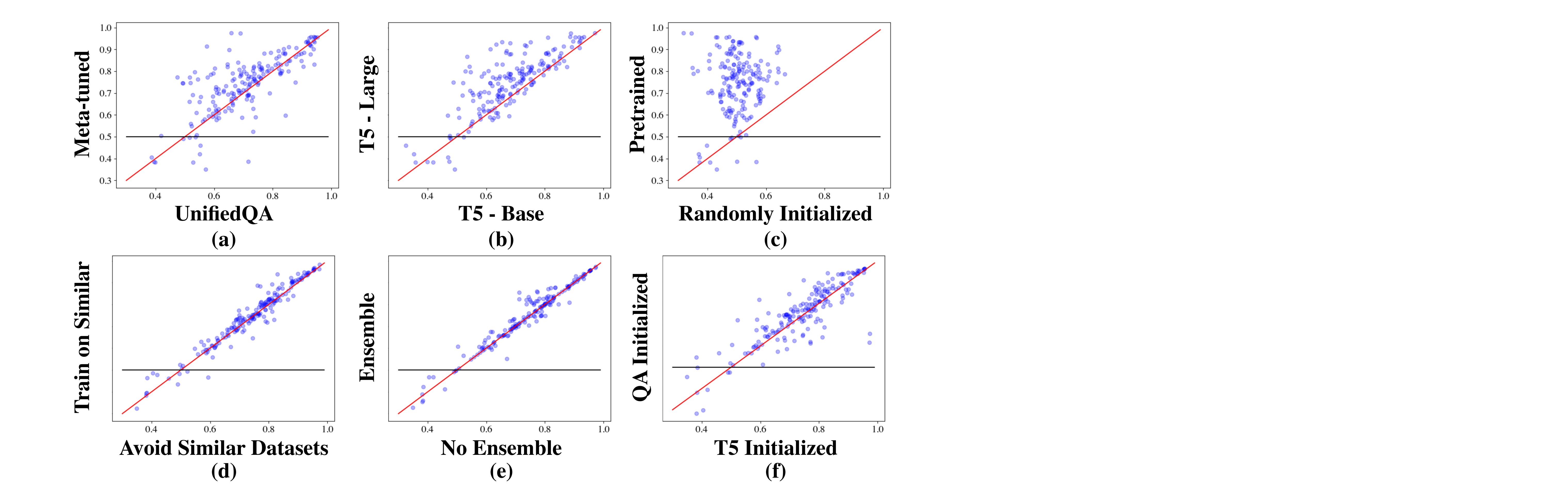}
    \caption{
    The interpretation of these figures can be seen in Figure \ref{fig:metrics-explained}.
    (a) compares a meta-tuned model ($y$) against UnifiedQA ($x$); (b) compares T5-Large (770 M parameters) against T5-base (220M); (c) compares the T5 pre-trained initialization against the random initialization; (d), (e), and (f) investigate whether performance can be improved by training on similar datasets, ensembling different label descriptions (questions), and initializing with UnifiedQA. \textbf{Conclusion}:
    Since most dots are above the red line $y=x$ for all 6 figures and above the random guess baseline ($y=0.5$) by a robust margin, all conclusions listed at the beginning of Section \ref{sec:results} hold.}
    \label{fig:main_large}
\end{figure*}

We investigate and validate the following hypotheses, sorted by importance in descending order. 

\begin{itemize}
    \item Meta-tuned models outperform general question answering models in zero-shot classification.
    \item Larger pre-trained models are better.
    \item Pre-training does the heavy lifting.
    \item Performance can be improved by training on similar datasets, initializing with a QA model, or ensembling label descriptions.
    \item Early stopping is crucial to performance.
\end{itemize}

\paragraph{Meta-tuned models are better.} 
We compare a meta-tuned T5-Large model (770 M parameters)\footnote{This model is initialized with T5, not UnifiedQA.} with the same-sized UnifiedQA model \cite{khashabi-etal-2020-unifiedqa} out of the box. 
Relevant descriptive statistics can be seen in the first row of Table \ref{tab:main-tab} and Figure \ref{fig:main_large} (a). 
Adapting the model for ZSC improves the average AUC-ROC by 3.3\%. 
\paragraph{Larger pre-trained models are better.}
We compare T5-Base (220 Million parameters) against T5-Large (770 M).
The statistics can be seen in the second row of Table \ref{tab:main-tab} and Figure \ref{fig:main_large} (b). 
Increasing the model size from 220 M to 770M improves the average AUC-ROC by 6.3\%. 

\paragraph{Pre-training does the heavy lifting.} In Figure (c) and the third row of Table \ref{tab:main-tab}, we compare pre-trained and random initializations, where the latter cannot beat the random baseline (average AUC-ROC 0.503).
Hence, meta-tuning alone is far from enabling the model to perform ZSC.
An intuitive interpretation is that the model already ``knows" how to perform ZSC after pre-training under the LM objective, and learns how to use this knowledge during meta-tuning.

\paragraph{Training on similar datasets improves performance.}
Unlike before, we no longer avoid training on similar datasets from the same group. 
Instead, we perform straightforward leave-one-out cross-validation.
The statistics can be seen in the fourth row of Table \ref{tab:main-tab} and Figure \ref{fig:main_large} (d), and it improves the average AUC-ROC by 0.7\%.
The performance gain is not as significant as increasing the model size or adapting for ZSC.
We conjecture that it is because we have not collected enough datasets; otherwise, there might be more similar datasets, hence improving ZSC performance.

\paragraph{Ensembling label descriptions improves performance.}
Instead of asking the model a single question for each label and obtain the probability of the answer being ``\textit{Yes}", we can average the probability obtained by asking multiple questions with the same meaning. 
This approach is different from traditional ensembling, which typically needs to store/train multiple models to average across them. 
The fifth row of Table \ref{tab:main-tab} and Figure \ref{fig:main_large} (e) verifies that ensembling descriptions improves performance slightly (0.7\% AUC-ROC score).

\paragraph{Initializing with UnifiedQA improves performance.}
Figure \ref{fig:main_large} (f) and the sixth row of Table \ref{tab:main-tab} compare the UnifiedQA against against the T5 initialization.
Initializing with UnifiedQA improves average AUC-ROC by 1.1\%.

\begin{figure}[t!]
    \centering
    \includegraphics[width=\columnwidth]{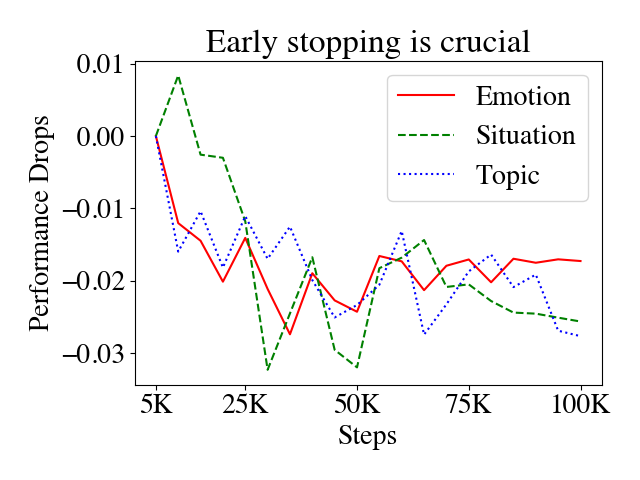}
    \caption{Each curve corresponds to the models' performance on a dataset from \citet{yin-etal-2019-benchmarking}. $x$-value is the number of training steps; $y$-value is the average AUC-ROC score across all label descriptions, relative to the value at step 5000. Training for too long decreases performance on unseen tasks.}
    \label{fig:longrunning}
\end{figure}

\paragraph{Early stopping is crucial to performance.}
If we train the model for too long, the model might simply ``memorize" that certain label descriptions correspond to certain training tasks, and the performance on unseen tasks may drop. 
To explore this possibility, we meta-tune our models for 100K steps, which is 20 times as long as our default setting and encourages the model to memorize the training tasks. 
We then evaluate them on the three benchmark zero-shot classification datasets by \citet{yin-etal-2019-benchmarking} (which we describe in more details in the next section).
We calculate the average AUC-ROC across all label descriptions for each of the 3 datasets, and plot them in Figure \ref{fig:longrunning}.

The performance decreases \footnote{Kendall rank correlation coefficients are negative with $p < 0.005$ for topic and situation classification} as training continues. 
On the other hand, however, the performance drop of 3\% in AUC-ROC is not fatal and the model's performance is still much better than random guesses.
%We conjecture that ``memorization" will lead to even less performance decrease when there are more different label descriptions in the training data. 

\subsection{Robustness Checks} \label{sec:robust}
We examine a series of additional results to make sure our conclusions are robust.
The observed improvements in Table \ref{tab:main-tab} and Figure \ref{fig:main_large} might be caused by the improvement of a small number of labels that are annotated with more descriptions, or by the improvement on a dataset with more distinct labels.
Appendix \ref{appendix-sec:other-uniform} compares the performance by assigning equal weights to each label/datasets.

To provide additional supporting evidence for our forecast that larger models are better, Appendix \ref{appendix-sec:t5-larger-better} compares a 60M-parameter model against a 220M-parameter model, and finds that the latter is much better.
One concern, however, is that our models are initialized with T5 \cite{raffel2019exploring}, which is trained on the open web and might have seen the datasets we gathered. 
Therefore, larger models might be better simply because they are better at memorization \cite{sagawa2020investigation}.
Appendix \ref{appendix-sec:bert-larger-better} addresses this by showing that larger models are also better with BERT initialization \cite{devlin-etal-2019-bert}, which is trained on Wikipedia and Book Corpus \cite{zhu2015aligning}.

We also report the models' performance on each dataset for readers' reference in Appendix \ref{appendix-sec:breakdown}.

\subsection{Comparison with \citet{yin-etal-2019-benchmarking}} \label{sec:comparison}

\begin{table}[]
    \centering
    \begin{tabular}{lccc}
        \hline
        Model & emotion & situation & topic \\
        \hline
        \citet{yin-etal-2019-benchmarking} & 25.2 & 38.0 & 52.1\\
        Meta-tuned & \textbf{28.2} & \textbf{48.4} & \textbf{54.3 }\\
        \hline
    \end{tabular}
    \caption{``Prior" means the best performing system from \citet{yin-etal-2019-benchmarking} for each dataset; ``Meta-tuned" means meta-tuning on RoBERTa.
    Our approach is better on all three datasets.
    }
    \label{tab:benchmarking}
\end{table}

This section shows that our approach has higher performance than the zero-shot classification system built by \citet{yin-etal-2019-benchmarking}.
Their system ensembles several natural language inference models based on RoBERTA-Large (355M parameters, \citet{liu2020roberta}), and another model trained to categorize Wikipedia articles.
It was evaluated on three classification datasets:

\begin{itemize}
    \item topic (10-way): classifies article domains, such as \textit{family \& relationship, education, sports}, etc. The metric is accuracy. 
    \item emotion (10-way): classifies emotion types, such as \textit{joy, anger, guilt, shame}, etc. The metric is label-weighted F1.
    \item situation (12-way): classifies disaster situations, e.g. \textit{regime change, crime \& violence}, and the resource they need, e.g. \textit{search \& rescue}. The metric is label-weighted F1.
\end{itemize}

We use the exact same evaluation metrics as in \citet{yin-etal-2019-benchmarking}, and the same label resolution strategy when the model answers ``Yes"\footnote{or ``Entailment" for natural language inference models.} for multi-label classification. 
Concretely, when the model predicts ``Yes" on multiple labels, the one with the highest probability is selected. 
For a fair comparison, we meta-tune RoBERTa of the same size and compare it with the highest performing model in \citet{yin-etal-2019-benchmarking} for each of the three datasets.

The results are in Table \ref{tab:benchmarking}, and our model has higher performance across all 3 datasets using the same pre-training method.

\section{Discussion and Future Directions} \label{sec:implications}

\paragraph{Main takeaways} 
We construct a dataset of classification datasets to adapt the language model for zero-shot classification via meta-tuning.
The adapted model outperforms a general-purpose question answering model and the prior state of the art based on natural language inference.
We forecast that meta-tuning would be more effective on larger models, and the current engineering ceiling for zero-shot learning might have been broadly under-estimated.

\paragraph{Aggregating and unifying datasets}
The main bottleneck of our research is to manually gather a wide range of datasets and unify their format.
The difficulties are:
1) we need to brainstorm and review the NLP literature extensively to decide what new tasks to look for; 
2) different datasets encode their data in different formats, and we need to write programs manually for each of them to convert to the desired format; 
3) it is hard to tell the quality of a dataset purely by its provenance, and sometimes we need to examine the dataset manually. 
If we as a community can aggregate and unify datasets better, we could potentially train and evaluate zero-shot learning models at a larger scale.
%Our research would have been much easier if there is a platform that gathers all the NLP classification datasets, unifies them into a single format, and provides information about their individual quality.

\paragraph{Meta-tuning as a probe}
There is a growing interest in measuring the intelligence \cite{hendrycks2021aligning, hendrycks2021measuring} or the few-shot learning ability \cite{brown2020language} of large language models like GPT-3. 
However, since these models are not adapted to answer those prompts \cite{DBLP:journals/corr/abs-2104-08315}, we suspect that its knowledge and true potential to perform few-shot learning is much higher than reported.
Since pre-training does the heavy lifting and meta-tuning is unlikely to provide additional ZSC ability to the model,
we can potentially first use meta-tuning as a probe to make them adapted to answering prompts before measuring their performance.

Still, to make this methodology rigorous, interpreting and controlling the strength of the probes will be an important future direction \cite{hewitt-liang-2019-designing}.
For example, 
%if the meta-tuning procedure becomes so strong in the future that it enables a randomly initialized model to answer prompts correctly, or 
if the training set contains a prompt that is too similar to the prompt to be tested, the probe will be meaningless.

\paragraph{Beyond Shallow Correlations}
One possibility is that the model only learns shallow statistical correlations from meta-tuning rather than ``more sophisticated reasoning skills".
For example, the word ``exciting" might occur in positive reviews more. 
This is unlikely, given that larger models are consistently better than smaller or randomly initialized ones.
To explain this performance gap, larger models must have learned to use more complicated features during meta-tuning.

\paragraph{Relation to Meta/Multitask-Learning} 
Our method is closely related to, but different from meta-learning \cite{yin2020meta, murtydreca} and multi-task learning \cite{DBLP:journals/corr/abs-2104-08835, aghajanyan2021muppet}. 
Both meta-learning and multitask-learning typically involve at least a few examples from the target task; in our setup, however, the model does not learn from any target task examples. The “meta” in our name does not mean “meta-learning”, but reflects the fact that our model learns from a meta-dataset of tasks. 

Nevertheless, our framework can be easily adapted to a few-shot learning setup, which enables the language model to learn to learn from in-context examples (see below). 
Since this approach models the learning process as a sequence classification problem, it can be seen as a form of meta-learning similar to \cite{ravi2016optimization}.

\paragraph{Annotating Prompts} 
Three of our authors annotated the label descriptions.
Since they are all Computer Science major students who understand machine learning and natural language processing, they might not be representative of the final user population of this ZSC application.
Annotating prompts that match the target user distribution will be an important research direction. 

Additionally, shorter and more natural descriptions sometimes fail to capture the exact semantics of the label. 
For example, in \citet{yin-etal-2019-benchmarking}, the description of the label ``medical" is ``\textit{people need medical assistance}"; or alternatively, it can be longer but more accurate: ``\textit{people need an allied health professional who supports the work of physicians and other health professionals}".
How to scalably generate more accurate and detailed label descriptions without expert efforts will be another future direction.

\paragraph{Optimizing Prompts}
Our work is complementary to recent works that optimize the prompts to achieve better accuracy.
Even if our meta-tuned model is specialized in answering prompts, it might still react very differently towards different prompts.
For example, in the stance classification dataset \cite{barbieri-etal-2020-tweeteval}, we annotated two label descriptions (prompts) for the same label: ``Does this post support atheism?" and ``Is the post against having religious beliefs?".
They have similar meanings, but the former has much lower accuracy than the later.
We conjecture that this is because the model cannot ground abstract concepts like ``atheism".

\paragraph{Other extensions}
We conjecture that meta-tuning can be extended to more diverse tasks beyond zero-shot binary classification. 
To extend to multi-label classification, we need to develop a procedure to resolve the labels when the model predicts positive for more than one labels.
To extend to few-shot learning, we need to increase the context length to fit several training examples into the input, which requires a larger context window and hence more computational resources.
%For example, even 2-shot learning, which includes one positive and one negative example, would triple the context size, and hence increasing the computation requirement by 8 times. 
To extend to other sequence generation tasks, we need to collect a wide range of diverse sequence generation tasks to meta-tune the model, such as machine translation, summarization, free-form question answering, grammar correction, etc.

\section*{Acknowledgements}
We thank Eric Wallace for his feedbacks throughout the project. 
We thank Steven Cao, David Gaddy, Haizhi Lai, Jacob Steinhardt, Kevin Yang and anonymous reviewers for their comments on the paper.

\bibliography{anthology,custom}
\bibliographystyle{acl_natbib}

\appendix
\pagebreak
\section{Ethics} \label{sec:ethics}

\paragraph{Data and incentives} 
In the existing prompting framework, end users send the natural language descriptions and a few training examples to the large language model inference API to perform few-shot learning \cite{brown2020language}.
This becomes a natural source of training data for meta-tuning.
Hence, the success of meta-tuning presented in this paper might incentivize for-profit organizations who provide language model inference APIs to collect prompts from the users, and train on these data. 

\paragraph{Privacy, security, and fairness} 
If a model is meta-tuned on user-provided data, certain security, privacy and fairness concerns can potentially emerge. 
For example, \citet{carlini2020extracting} shows that it is possible to extract the training data from large language models, and hence meta-tuned systems might expose some users' prompts to other users. 
\citet{wallace2020customizing} shows that it is possible to poison the model through training data and trigger unwanted behaviors; 
the meta-tuning procedure might be susceptible to these data poisoning attacks as well.
Finally, meta-tuning might perpetuate existing societal biases hidden in the users' prompts  \cite{NIPS2016_a486cd07}. 

If not addressed properly, these concerns might have a broader negative societal impact through meta-tuning.
Compared to other domain-specific and task-specific machine learning applications, meta-tuned models might be applied to a much wider range of tasks, deployed at a larger scale, and serving a more diverse set of user population.
Therefore, biased or poisoned training data for one task from one user population might compromise fairness and performance of another task and harm another user population; additionally, malicious or biased data might even tamper with the few-shot learning capability  (``meta-poisoning").

\paragraph{Potential abuse} 
As shown in Figure \ref{fig:main_large}, the AUC-ROC score for a lot of tasks are still well below 0.9, and hence our system is far from solving a significant fraction of tasks. 
Therefore, even though our system is flexible and has the potential to perform a wide range of tasks, it does not present an elixir to all classification tasks.
Particularly, it should not be applied to higher stake scenarios (e.g. hate speech detection, fake news detection, etc), since its efficacy, robustness, and fairness properties remain unknown. 

\section{Datasets} \label{appendix-sec:dataset}

\paragraph{IMDB movie review sentiment classification} \cite{maas-etal-2011-learning}.
Classifies whether the user likes the movie.

\textsc{Positive}: ``'My favourite police series of all time turns to a TV-film. Does it work? Yes. ..."

\textsc{Negative}:  `` "Stupid! Stupid! Stupid! I can not stand Ben stiller anymore."

\paragraph{Zero Shot Emotion Classification} \cite{yin-etal-2019-benchmarking}.
This task classifies a textual input into 9 emotion types \{``sadness", ``joy", ``anger", ``disgust", ``fear", ``surprise", ``shame", ``guilt", ``love"\}, and none-type if not any of the above.
For example,

\textsc{Joy}: ``\textit{Making new friends is always fun , specially when playing dress up}"

\textsc{Anger}: ``\textit{People that smoke cigarettes irritate my soul.}"

\paragraph{Zero Shot topic Classification} \cite{yin-etal-2019-benchmarking}.
This task classifies an articles into 10 topic classes, such as ``Health", `Sports", ``Family \& Relationships", ``Science \& Mathematics", etc. 

\textsc{Science}: ``\textit{How the human species evolved ? How the human species evolved ? A tough question as it overlaps science and theology ...}"

\textsc{Sports}: ``\textit{Number one formula one driver ? Depends on your question.Kimi Raikkonen won more races ...}"

\paragraph{Zero Shot Situation Classification} \cite{yin-etal-2019-benchmarking}. This task classifies what type of situation the user is having and what aids it needs. 
The classes include ``food", ``terrorism", etc.

\textsc{Food}: "livestock , crops , rural and urban housing , roads , river and canal embankments , and other infrastructure suffered extensive damage"

\textsc{Terrorism}: "Daish attacked Tahurmahu village with chemical gas."

\paragraph{}

% Kristy Datasets 
\paragraph{SemEval 2017 Task 4}
\cite{barbieri-etal-2020-tweeteval} \cite{rosenthal-etal-2017-semeval}. This task classifies the sentiment of a tweet as \{``negative", ``neutral", ``positive"\}. For example,

\textsc{Negative}: ``\textit{That sucks if you have to take the SATs tomorrow}"

\textsc{Neutral}: ``\textit{@user Well said on HMW. Can you now address why Texans fans file out of the stadium midway through the 4th qtr of every game?}"

% Note: The example had quotes inside of them " " --> the example specifically was "This is the last Sunday w/o football .....,NFL is back baby". Might consider getting rid of the quotes?
\textsc{Positive}: ``\textit{"This is the last Sunday w/o football .....,NFL is back baby"}"

\paragraph{ETHICS Justice}
\cite{2020arXiv200802275H}. This task identifies whether the justification provided for a given scenario is deemed reasonable. For example, 
% Note: Reasonable and Unreasonable are not explicit labels that the dataset assigned, but rather the ones I created to categorize the examples

\textsc{Reasonable}: ``\textit{I deserve to be paid by my boss because I keep her house clean daily.}"

\textsc{Unreasonable}: ``\textit{I deserve to be given a brand new car by my boss because I keep her house clean daily.}"

\paragraph{ETHICS Deontology}
\cite{2020arXiv200802275H}. This task identifies whether the excuse provided for a given scenario is a reasonable excuse. For example, 

\textsc{Scenario}: ``\textit{Aren't you supposed to be preparing for the meeting?}"

\textsc{Excuse (Reasonable)}: ``\textit{No because I have everything already set up in the conference room.}"

\textsc{Excuse (Unreasonable)}: ``\textit{No because you don't have a meeting}"

\paragraph{ETHICS Common Sense}
\cite{2020arXiv200802275H}. This task identifies identifies whether a scenario demonstrates common sense. For example, 

%  Note: Common Sense and Not Common Sense are not explicit labels that the dataset assigned, but rather the ones I created to categorize the examples
\textsc{Common Sense}: ``\textit{I went to the principal's office to change my records before going to a different school.}"

\textsc{Not Common Sense}: ``\textit{I secured the loan because I would make the payments.}"

\paragraph{EURLEX57K}
\cite{chalkidis-etal-2019-large}. This task classifies European legislation. 

\textsc{National Currency}: ``\textit{Council Regulation (EC) No 2595/2000 of 27 November 2000 amending Regulation (EC) No 1103/97 on certain provisions relating to the introduction of the euro}"

\textsc{Southern Africa}: ``\textit{95/458/EC: Commission Regulation (EC) No 302/2006 of  20 February 2006  on import licences in respect of beef and veal products originating in Botswana, Kenya, Madagascar, Swaziland, Zimbabwe and Namibia}"

\paragraph{SemEval 2019 Task 6}
\cite{barbieri-etal-2020-tweeteval} \cite{zampieri-etal-2019-semeval}. This task classifies the tweet as either offensive or not offensive. For example, 

\textsc{Offensive}: ``\textit{@user She has become a parody unto herself? She has certainly taken some heat for being such an....well idiot. Could be optic too  Who know with Liberals  They're all optics.  No substance}"

\textsc{Not Offensive}: ``\textit{@user @user She is great. Hi Fiona!}"

\paragraph{Click Bait Detection}\footnote{\url{https://www.kaggle.com/c/clickbait-news-detection}}
This task detects whether a news title is a click bait.

\textsc{Click Bait}: ``\textit{Can You Pass This Basic Trigonometry Quiz}"

\textsc{Non Click Bait}: ``\textit{NASCAR driver Kyle Busch wins 2011 Jeff Byrd 500}".

\paragraph{Abstract Domain Classification}\footnote{\url{https://www.kaggle.com/abisheksudarshan/topic-modeling-for-research-articles?select=Train.csv}}
This classifies the abstract into 4 domains: ``Physcis", ``Maths", ``Computer Science", ``Statistics".
For example, 

\textsc{Physics}: ``\textit{a ever-growing datasets inside observational astronomy have challenged scientists inside many aspects, including an efficient and interactive data exploration and visualization. many tools have been developed to confront this challenge ...}"

\textsc{Maths}: ``\textit{a main result of this note was a existence of martingale solutions to a stochastic heat equation (she) inside the riemannian manifold ...}"

\paragraph{SemEval 2019 Task 5}
\cite{barbieri-etal-2020-tweeteval} \cite{basile-etal-2019-semeval}. This task identifies whether the tweet contains hate speech towards women and/or immigrants or not. For example, 

\textsc{Hate Speech}: ``\textit{This account was temporarily inactive due to an irrational woman reporting us to Twitter. What a lack of judgement, shocking. \#YesAllMen}"

\textsc{No Hate Speech}: ``\textit{@user nice new signage. Are you not concerned by Beatlemania -style hysterical crowds crongregating on you…}"

\paragraph{SemEval 2019 Task 8}
\cite{mihaylova-etal-2019-semeval}. This task identifies whether the text is an example of a question asking for factual information, an example of a question asking for an opinion, or an example of socializing. For example, 

\textsc{Factual}: ``\textit{is there any place i can find scented massage oils in qatar?}"

\textsc{Opinion}: ``\textit{hi there; i can see a lot of massage center here; but i dont which one is better. can someone help me which massage center is good...and how much will it cost me? thanks}"

\textsc{Socializing}: ``\textit{Hello people...let's play this game...you have to write something good about the person whose 'post' is above you on QL.You can write anything and you can write\&\#160; multiple times.}"

\paragraph{SemEval 2018 Task 3}
\cite{barbieri-etal-2020-tweeteval} \cite{van-hee-etal-2018-semeval}. This task identifies whether the tweet contains irony or not. For example, 

\textsc{Irony}: ``\textit{seeing ppl walking w/ crutches makes me really excited for the next 3 weeks of my life}"

\textsc{No Irony}: ``\textit{@user on stage at \#flzjingleball at the @user in \#Tampa \#iheartradio}"

\paragraph{SemEval 2018 Task 1}
\cite{barbieri-etal-2020-tweeteval, mohammad-etal-2018-semeval} This task classifies a tweet as one of 4 emotion types \{``sadness", ``joy", ``anger", ``optimism"\}. For example,

\textsc{Sadness}: ``\textit{@user I so wish you could someday come to Spain with the play, I can't believe I'm not going to see it \#sad}"

\textsc{Joy}: ``\textit{\#ThisIsUs has messed with my mind \&amp; now I'm anticipating the next episode with \#apprehension \&amp; \#delight! \#isthereahelplineforthis}"

\textsc{Anger}: ``\textit{@user Haters!!! You are low in self worth. Self righteous in your delusions. You cower at the thought of change. Change is inevitable.}"

\textsc{Optimism}: ``\textit{Don't be \#afraid of the space between your \#dreams and \#reality. If you can \#dream it, you can \#make it so}"

\paragraph{SemEval 2016 Task 6}
\cite{mohammad-etal-2016-semeval, barbieri-etal-2020-tweeteval} This task classifies a tweet's stance as \{``neutral", ``against", ``favor"\}. Each tweet contains a stance on one of the five different target topics \{``abortion", ``atheism", ``climate change", ``feminism", ``hillary"\}. For example, 

\textsc{Neutral}: ``\textit{@user maybe that's what he wants \#SemST}"

\textsc{Against}: ``\textit{Life is \#precious \& so are babies, mothers, \& fathers. Please support the sanctity of Human Life. Think \#SemST}"

\textsc{Favour}: ``\textit{@user @user Nothing to do with me.  It's not my choice, nor is it yours, to dictate what another woman chooses. \#feminism \#SemST}"

\paragraph{SemEval 2020 Task 6} \cite{spala-etal-2020-semeval}.
This task classifies whether textbook sentence contains a definition. For example,

\textsc{Contains Definition}: ``\textit{Since 2005, automated sequencing techniques used by laboratories are under the umbrella of next-generation sequencing, which is a group of automated techniques used for rapid DNA sequencing}"

\textsc{Doesn't Contain Definition}: ``\textit{These automated low-cost sequencers can generate sequences of hundreds of thousands or millions of short fragments (25 to 500 base pairs ) in the span of one day.}"

\paragraph{TREC} \cite{li-roth-2002-learning}.
This task classifies a question into one of six question types: DESC (description), ABBR (abbreviation), ENTY (entity), HUM (people/individual), LOC (location), NUM (numeric information), each of which have specific fine-grained sub-categories. For example,

\textsc{DESC}: ``\textit{How did serfdom develop in and then leave Russia?}"

\textsc{ABBR}: ``\textit{What is the full form of .com?}"

\textsc{ENTY}: ``\textit{What films featured the character Popeye Doyle?}"

\textsc{HUM}: ``\textit{What contemptible scoundrel stole the cork from my lunch?}"

\textsc{LOC}: ``\textit{What sprawling U.S. state boasts the most airports?}"

\textsc{NUM}: ``\textit{How many Jews were executed in concentration camps during WWII?}"

\paragraph{SUBJ} \cite{pang-lee-2004-sentimental}.
This task classifies a sentence as being subjective or objective. For example, 

\textsc{Subjective}: ``\textit{smart and alert, thirteen conversations about one thing is a small gem.}"

\textsc{Objective}: ``\textit{the movie begins in the past where a young boy named sam attempts to save celebi from a hunter.}"

% Jacky Datasets 

\paragraph{The Corpus of Linguistic Acceptability} 
\cite{warstadt2018neural}.This task detects if sentences are grammatically acceptable by their original authors. For example,

\textsc{Grammatically Acceptable}: ``\textit{Her little sister will disagree with her.}"

\textsc{Grammatically Not Acceptable}: ``\textit{Has not Henri studied for his exam?}"

\paragraph{The Multi-Genre NLI Corpus} \cite{williams-etal-2018-broad}.
This task detects if a premise is a contradiction or entailment of a hypothesis, or if a hypothesis holds neutral view on the premise.. For example,

\textsc{Neutral}: ``\textit{Premise: Exoatmospheric Kill Vehicles orbiting Earth would be programmed to collide with warheads. Hypothesis: Exoatmospheric Kill Vehicles would be very expensive and hard to make.}"

\textsc{Entailment}: ``\textit{Premise: so we have to run our clocks up forward an hour and i sure do hate to loose that hour of sleep in the morning. Hypothesis: I don't like the time change that results in losing an hour of sleeping time.}"

\textsc{Contradiction}: ``\textit{Premise: The mayor originally hoped groundbreaking would take place six months ago, but it hasn't happened yet. Hypothesis: The mayor doesn't want groundbreaking to happen at all.}"

\paragraph{Metaphor as a Medium for Emotion: An Empirical Study} \cite{MohammadST16}.
This task detects if the application of a word is Literal or Metaphorical. For example,

\textsc{Word: abuse} 

\textsc{Literal: }``\textit{This boss abuses his workers.}"

\textsc{Metaphorical: }``\textit{Her husband often abuses alcohol.}"

\paragraph{Political Preference Classification} 
\cite{allaway-mckeown-2020-zero}.
This task predicts a comment's stand point on a political topic. For example,

\textsc{Topic: Companies Regulation} 

\textsc{Con: }``\textit{Regulation of corporations has been subverted by corporations. States that incorporate corporations are not equipped to regulate corporations that are rich enough to influence elections, are rich enough to muster a legal team that can bankrupt the state. Money from corporations and their principals cannot be permitted in the political process if democracy is to survive.}"

\textsc{Pro: }``\textit{Regulation is to a corporation what a conscience is to a living person. Without a conscience, we would all be sociopaths. Corporations do not have a conscience, thus they need regulation to make sure they are focused on benefiting society instead on merely benefiting themselves.}"

\textsc{Neutral: }``\textit{Without government to ensure their behavior, companies will attempt to make a profit even to the DETRIMENT of the society that supports the business. We have seen this in the environment, in finances, in their treatment of workers and customers. Enough.}"

\paragraph{Airline Service Review}\footnote{\url{https://www.kaggle.com/welkin10/airline-sentiment}}
This task classifies if an airline review has a positive or negative sentiment. For example,

\textsc{Positive: }``\textit{This is such a great deal! Already thinking about my 2nd trip to Australia; I haven't even gone on my 1st trip yet!}"

\textsc{Negative: }``\textit{amazing to me that we can't get any cold air from the vents.}"

\paragraph{Covid-19 Tweets Sentiment Analysis}
\footnote{\url{https://www.kaggle.com/datatattle/covid-19-nlp-text-classification?select=Corona\_NLP\_test.csv}}
This task classifies if a tweet has a positive or negative sentiment. For example,

\textsc{Positive: }``\textit{Taken by Henk Zwoferink on Saturday in Wargl, our black beauty hauled a train bringing the last tourists home. Our colleagues are \#workinghard to keep supply chains running while respecting the measures to ensure everyone's \#safety. A pleasure to work with such \#DedicatedPeople!}"

\textsc{Negative: }``\textit{So far, the Minister does not seem to have made statement on the catastrophe that can develop if the issue of markets operation is not addressed. Food insecurity has potential to make current Covid-19 panic look like a kindergarten and could lead to riots. I submit.}"

    \paragraph{Hotel Review}\footnote{\url{https://www.kaggle.com/andrewmvd/trip-advisor-hotel-reviews}}
This task predicts if a hotel review is a positive or negative review. For example,

\textsc{Negative: }``\textit{The single rooms like hospital rooms single rooms hotel sparse intentional know ugly like trapped hospital white walls sink basin room small rectangle shape.the beds hard rocks blankets rough really noisy.this overrated hotel stayed fans type hotels}"

\textsc{Positive: }``\textit{loved stay, stayed univ, inn 10 days april 2005 thoroughly enjoyed, free parking clean spacious room friendly staff great breakfast snack, loved location, definitely stay,  }"

\paragraph{Stock Market Sentiment}\footnote{\url{https://www.kaggle.com/yash612/stockmarket-sentiment-dataset}}
This task predicts if a comment holds a positive or negative view on the performance of the stock market. For example,

\textsc{Negative: }``\textit{GPS wow that wa s a fast fast fade...}"

\textsc{Positive: }``\textit{user Maykiljil posted that: I agree that MSFT is going higher \& possibly north of 30}"

\paragraph{AG-News} \cite{10.5555/2969239.2969312}.
This task classifies the topic of news based on their contents. For example,

\textsc{World News: }``\textit{Greek duo could miss drugs hearing}"

\textsc{Sports News: }``\textit{AL Wrap: Olerud Cheers Yankees by Sinking Ex-Team}"

\textsc{Business News: }``\textit{Lowe's Second-Quarter Profit Rises}"

\textsc{Tech News: }``\textit{Satellite boosts Olympic security}"

\paragraph{Real and Fake News}\footnote{\url{https://www.kaggle.com/amananandrai/ag-news-classification-dataset?select=train.csv}}
This task classifies if a news is fake or real. For example,

\textsc{Real: }``\textit{WASHINGTON (Reuters) - Alabama Secretary of State John Merrill said he will certify Democratic Senator-elect Doug Jones as winner on Thursday despite opponent Roy Mooreâ\\x80\\x99s challenge, in a phone call on CNN. Moore, a conservative who had faced allegations of groping teenage girls when he was in his 30s, filed a court challenge late on Wednesday to the outcome of a U.S. Senate election he unexpectedly lost.}"

\textsc{Fake: }``\textit{Ronald Reagan shut down the Berkeley protests many years ago THIS is how you do it!}"

\paragraph{Disaster Tweets}\footnote{\url{https://www.kaggle.com/c/nlp-getting-started/data?select=train.csv}}
This task detects if a tweet announces an emergency or a disaster. For example,

\textsc{Contains Disaster: }``\textit{Our Deeds are the Reason of this \#earthquake May ALLAH Forgive us all.}"

\textsc{Does not Contain Disaster: }``\textit{My dog attacked me for my food \#pugprobs.}"

\paragraph{Obama vs Trump Tweets}\footnote{\url{https://www.kaggle.com/shaharz/classifying-tweets-of-trump-and-obama}}
This task detects if a tweet was send by Obama or Trump. For example,

\textsc{Obama: }``\textit{Michelle and I are delighted to congratulate Prince Harry and Meghan Markle on their engagement. We wish you a lifetime of joy and happiness together.}"

\textsc{Trump: }``\textit{Together, we dream of a Korea that is free, a peninsula that is safe, and families that are reunited once again!}"

\paragraph{Kaggle Sexually Explicit Tweets}\footnote{\url{https://www.kaggle.com/harsh03/sexually-explicit-comments}}
This dataset provides positive examples of profane comments. For example,

\textsc{Explicit}``\textit{What do guys say when you get naked in front of them for the first time?}"

\paragraph{Democratic vs Republican Tweets}\footnote{\url{https://www.kaggle.com/kapastor/democratvsrepublicantweets?select=ExtractedTweets.csv}}
This task detects if a tweet was send by the Democratic or Republican Party. For example,

\textsc{Democratic: }``\textit{\#YuccaMountain would require moving tens of thousands of metric tons of radioactive waste across the country and through Southern Nevada.}"

\textsc{Republican: }``\textit{Stopped by One Hour Heating\&amp; Air Conditioning to discuss the benefits tax reform will bring to their business.}"

\paragraph{Women E-commerce Clothing Reviews}\footnote{\url{https://www.kaggle.com/nicapotato/womens-ecommerce-clothing-reviews}}
This task predicts if the buyer likes or recommends a product base on its review. For example,

\textsc{Like: }``\textit{After reading the previous reviews, i ordered a size larger. i am so glad i did it! it fits perfectly! i am 5'4"/115/32dd and went with the s regular. so beautiful! i can't wait to wear it!}"

\textsc{Dislike: }``\textit{The zipper broke on this piece the first time i wore it. very disappointing since i love the design. I'm actually going to try to replace the zipper myself with something stronger, but annoying that it's come to that.}"

\paragraph{Quora Question Pairs}\footnote{\url{https://www.kaggle.com/c/quora-question-pairs/data}}
This task predicts if a pair of Quora question is asking for the same thing. For example,

\textsc{Same: }``\textit{Question 1: How many months does it take to gain knowledge in developing Android apps from scratch?; Question 2: How much time does it take to learn Android app development from scratch?}"

\textsc{Different: }``\textit{Question 1: How would you review the site Waveclues? ; Question 2: Is there a good pay for reviews site out there?}"

\paragraph{Headline Sarcasm Detection}
\href{https://www.kaggle.com/rmisra/news-headlines-dataset-for-sarcasm-detection?select=Sarcasm_Headlines_Dataset_v2.json}{This task} detects if is a news headline contains scarcasm. For example,

\textsc{Sarcasm: }``\textit{guy who just wiped out immediately claims he's fine}"

\textsc{No Sarcasm: }``\textit{Donald trump effigies burn across Mexico in Easter ritual}"

\paragraph{Company Account Tweets}
\footnote{\url{https://www.kaggle.com/thoughtvector/customer-support-on-twitter}}
This task detects whether the tweet is targeted towards a company account. For example,

\textsc{Yes: }``\textit{@VirginTrains Oh, that's nice. What are you doing about it? What are you targets next year?}"

\textsc{No: }``\textit{@115738 That's the best kind of trick-or-treating. All treats, my friend. -Becky}"

\paragraph{SMS Spam Detection} \cite{almeida2013towards}
This task detects whether the SMS is a spam message. For example,

\textsc{Spam: }``\textit{Thank you, winner notified by sms. Good Luck! No future marketing reply STOP to 84122 customer services 08450542832}"

\textsc{Ham: }``\textit{Lol great now I am getting hungry.}"

\paragraph{Clothing Fitness}\cite{misra2018decomposing} Checking whether the customer complains that the cloth is too small or too large.
%\footnote{We later noticed that this dataset is extremely noisy.}

\textsc{Small}: ``runs a bit small. wish it fit".

\textsc{Large}: ``too big".

\paragraph{Water Problem Topic Classification} \footnote{\url{https://www.kaggle.com/vbmokin/nlp-reports-news-classification?select=water_problem_nlp_en_for_Kaggle_100.csv}} Classifying the topic of a report on water problems. 
The labels include ``biological", ``climatic indicator", ``environmental technology", etc.
For example,

\textsc{Biological}: ``\textit{Mineralization of organic phosphorus in bottom sediments reaches 40–80\% and as we found out during the project implementation it intensified in autumn-winter period.}"

\textsc{climatic indicator}: ``\textit{The average amount of precipitation in the lower part of the basin makes 470 mm to 540 mm. The relative average annual air humidity makes 60-65\%}".

\textsc{environmental technology}: ``\textit{Most of wastewater treatment facilities require urgent modernization and reconstruction}".

\paragraph{Sexist Statement Detection}\footnote{\url{https://www.kaggle.com/dgrosz/sexist-workplace-statements}}
This task classifies whether the statement is sexist. For example,

\textsc{Sexist: }``\textit{It's impossible for a girl to be faithful.}"

\textsc{Non Sexist: }``\textit{Without strength, can we work to create wealth?}"

\paragraph{Movie Spoiler Detection} \cite{spoilerdataset}
\footnote{\url{https://www.kaggle.com/rmisra/imdb-spoiler-dataset?select=IMDB_reviews.json}}
This task classifies whether the movie review is a spoiler. For example,

\textsc{Spoiler: }``\textit{I must say that this movie was good but several things were left unsaid. For those who have seen the movie know what I am talking about but for those who haven't, I don't want to give spoilers. I was also impressed by Vin Diesel's acting skills. Overall I have to say it was a good movie filled with several twists and turns.}"

\textsc{Non Spoiler: }``\textit{The Great Wall amazes with its spectacular effects, both on screen and sound. Usually I do not appreciate 3D movies, but in this case I felt like it worth it.However, being honest, the storytelling and the story itself had its weaknesses. There were many logical lapses, and for me, many details are still waiting to be answered.On the other hand, expect decent acting especially from the main characters.All in all, The Great Wall is a solid popcorn-movie, but I expected a more elaborated unfolding of the legend it tells about.}"

\paragraph{News Summary/headline Topic Classification}
\footnote{\url{https://www.kaggle.com/rmisra/news-category-dataset}}
This task classifies the topic of the summary of a news. For example,

\textsc{Politics: }``\textit{City and state officials said they received little advance warning of the decision.}"

\textsc{Business: }``\textit{The streaming giant's third-quarter earnings were nothing like the Upside Down.}"

\iffalse
\paragraph{Sexist Statement Detection}\footnote{\url{https://www.kaggle.com/dgrosz/sexist-workplace-statements}}
This task classifies whether the statement is sexist. For example,

\textsc{Sexist: }``\textit{It's impossible for a girl to be faithful.}"

\textsc{Non Sexist: }``\textit{Without strength, can we work to create wealth?}"
\fi

\section{Dataset Property Tags} \label{appendix-sec:dataset-property}
Here we list all the dataset property tags (Section \ref{main-sec:data}).
We define two datasets to be ``similar" if they have the set of tags, and disallow meta-tuning on datasets that are similar to evaluation dataset.

\textit{social media}: whether the source is from social media (e.g. tweets).

\textit{social/political}: whether the task is highly related to political/social topics. Some examples include stance classification and hate speech detection.

\textit{topic classification}: whether the task classifies the topics of the input.

\textit{good vs. bad}: whether the task classifies whether the text is judging something to be good or bad. 

\textit{paper}: whether input text comes from a paper. 

\textit{review}: whether the input text is a review of a product (e.g. movie, hotel). 

\textit{questions}: whether the input texts are questions. Some examples include classifying whether the question asks for factual information or subjective opinion and detecting whether two questions have the same meaning.

\textit{emotion}: whether the task classifies certain emotion in the text, for example ``hate", ``surprise", ``joy", etc. 

Besides, we do not assign tags to datasets that we are confident to be different enough from other tasks (e.g. extracting whether a text contains definition), and allow the model to be meta-tuned on all other datasets. 

\section{List of Label Descriptions} \label{appendix-sec:label-descriptions}

\begin{figure*}
    \centering
    \includegraphics[width=\textwidth]{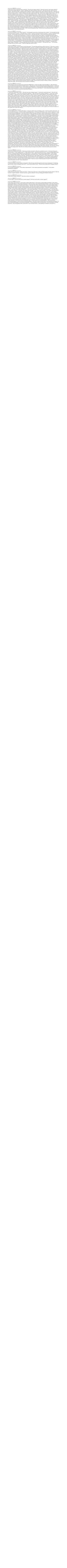}
    \caption{Label descriptions from the same group of datasets that are considered similar. ``*" at the beginning indicates that we are evaluating on this dataset.}
    \label{fig:desc1}
\end{figure*}

\begin{figure*}
    \centering
    \includegraphics[width=\textwidth]{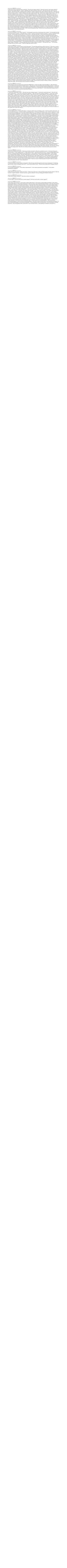}
    \caption{Label descriptions from the same group of datasets that are considered similar. ``*" at the beginning indicates that we are evaluating on this dataset.}
    \label{fig:desc2}
\end{figure*}

\begin{figure*}
    \centering
    \includegraphics[width=\textwidth]{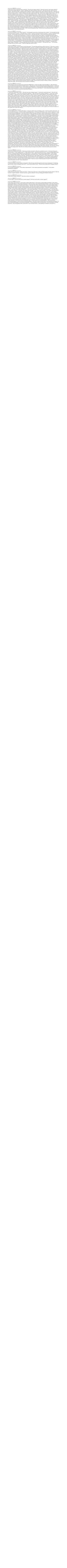}
    \caption{Label descriptions from the same group of datasets that are considered similar. ``*" at the beginning indicates that we are evaluating on this dataset.}
    \label{fig:desc3}
\end{figure*}

The comprehensive list of label descriptions and grouping can be seen in Figure \ref{fig:desc1}, \ref{fig:desc2}, and \ref{fig:desc3}.

\iffalse
Please refer to the appendix in our arXiv version: \url{https://arxiv.org/abs/2104.04670}. 
Somehow the acl\_pubcheck software package always gives us errors. 
\fi

\iffalse
\section{Comparing Different Sizes with BERT and T5}
\subsection{Larger T5 Models are Better} \label{appendix-sec:t5-larger-better}
Figure \ref{fig:comprehensive} (a) compares a meta-tuned model with 60 M parameters against one with 220M parameters, and (b) compares 220M parameters against 770M parameters.
For most label descriptions, larger models are better, and the ROC-AUC score is above the random guess baseline (above the black line) by a robust margin.

\subsection{Larger BERT Models are Better} \label{appendix-sec:bert-larger-better}
We concatenate the textual input to be classified and the label description (question) and feed them into the BERT model; to classify between ``Yes" and ``No", we add a linear layer on top of the pooled output (BERT-For-Sequence-Classification, \citet{wolf-etal-2020-transformers}).

Figure \ref{fig:comprehensive} (c) compares a meta-tuned model with 41 M parameters against one with 110M parameters, and (d) compares 110 parameters against 340M parameters.
For most label descriptions, larger models are better, and the ROC-AUC score is above the random guess baseline (above the black line) by a robust margin.
\fi

\section{Robustness Checks} \label{appendix-sec:comprehensive-results}
We report all the descriptive statistics mentioned in Section \ref{sec:metrics} under 3 different types of description weighting.
We additionally compare T5-small vs. T5-base, BERT-medium vs. BERT-Base and BERT-Base vs. BERT Large. 
All the results can be seen in Table \ref{tab:uniformlabeldescription}, \ref{tab:uniformlabel}, and \ref{tab:uniformdataset}
Due to space constraint, we abbreviate $\mathbb{P}[\Delta > t]$ as $> t$ if $t$ is positive, and $< t$ if $t$ is negative. 
Notice that, since we only have around 20 datasets to evaluate the model, most of the results presented here are not statistically significant at the dataset level;
nevertheless,  

\subsection{Different Description Weighting} \label{appendix-sec:other-uniform}
We weight each label and dataset equally in Table \ref{tab:uniformlabel} and \ref{tab:uniformdataset}.
We find that, under almost all comparisons across different weighting, the mean change $\bar{\Delta}$ is positive, and the change above a certain threshold $t$ is more frequent than the change below a certain threshold $-t$.
The only single exception the ``Ensemble" row in Table \ref{tab:uniformdataset}, where there are slightly more datasets where the change is lower than -1\%.
Nevertheless, given that the trend is still positive under $t = 5\%$ and $10\%$, and two other description weightings, we may still conclude that ensembling label descriptions is more likely to improve model performance.

\subsection{Larger T5 Models are Better} \label{appendix-sec:t5-larger-better}
In addition to comparing T5-Base (220 Million parameters) vs. T5-Large (770M), we also compare T5-small (60M) vs. T5-base (220M). 
Across all metrics, larger models are significantly better. 
Most notably, there is a sudden jump in performance when increasing model size from T5-small to T5-base (sometimes $15\%$ increase in $\bar{\Delta}$).

\subsection{Larger BERT Models are Better} \label{appendix-sec:bert-larger-better}
We also compare different sizes of BERT \cite{turc2019well} (41, 110, and 330M) parameters. 
Across all metrics, larger models are significantly better.

\begin{table*}[]
    \centering
    \begin{tabular}{lrrrrrrrr}
\hline
{} &  $\bar{\Delta}$&  $>$ 1\% &   $<$ -1\% &  $>$ 5\% &  $<$ -5\%  &  $>$ 10\% &  $<$-10\% &  $std(\Delta)$ \\
\hline
Meta-tuned vs QA &            3.3\% &            59.5\% &             28.1\% &            31.4\% &             10.3\% &           15.7\% &             5.9\% &        9.5\% \\
220 vs 770M (T5)      &            6.3\% &            75.1\% &             15.1\% &            47.6\% &              2.7\% &           27.0\% &             0.5\% &        8.1\% \\
Pre-trained vs. Random  & 23.8\% &          95.7\% &              3.2\% &            91.4\% &              1.6\% &           83.2\% &             1.1\% &       14.0\%\\
Ensemble                &            0.7\% &            28.9\% &             16.8\% &             8.7\% &              1.7\% &            1.7\% &             0.6\% &        3.1\% \\
Initialized with QA     &            1.1\% &            54.1\% &             24.3\% &            24.3\% &             11.9\% &            6.5\% &             4.9\% &        6.9\% \\
Train on similar        &            0.7\% &            43.8\% &             20.5\% &             6.5\% &              4.3\% &            1.6\% &             1.1\% &        3.2\% \\
60 vs 220M (T5)      &           14.4\% &            86.5\% &             10.3\% &            79.5\% &              4.3\% &           61.1\% &             2.2\% &       12.6\% \\
41 vs. 110M (BERT)   &            4.3\% &            65.9\% &             22.7\% &            40.0\% &             10.8\% &           20.5\% &             5.9\% &        9.1\% \\
110 vs. 340M (BERT)    &            1.4\% &            46.5\% &             35.7\% &            23.8\% &             17.3\% &           11.4\% &             6.5\% &        8.5\% \\
\hline
\end{tabular}
    \caption{All results, with metrics explained in Section \ref{sec:metrics} and Appendix \ref{appendix-sec:comprehensive-results}. Each label description is weighted equally.}
    \label{tab:uniformlabeldescription}
\end{table*}

\begin{table*}[]
    \centering
    \begin{tabular}{lrrrrrrrr}
\hline
{} &  $\bar{\Delta}$&  $>$ 1\% &   $<$ -1\% &  $>$ 5\% &  $<$ -5\%  &  $>$ 10\% &  $<$-10\% &  $std(\Delta)$ \\
\hline
Meta-tuned vs QA &            3.0\% &            57.5\% &             30.7\% &            31.3\% &             11.5\% &           16.2\% &             7.3\% &       10.2\% \\
220M vs 770M (T5)      &            5.8\% &            75.8\% &             15.5\% &            46.9\% &              3.5\% &           25.6\% &             1.4\% &        7.8\% \\
Pre-trained vs. Random  & 23.7\% &             93.5\% &              5.5\% &            89.4\% &              3.4\% &           82.5\% &             2.1\% &       15.1\% \\
Ensemble                &            0.5\% &            25.0\% &             18.8\% &             6.9\% &              1.6\% &            1.7\% &             0.7\% &        3.1\% \\
Initialized with QA     &            1.2\% &            54.0\% &             24.0\% &            26.0\% &             11.8\% &            8.1\% &             5.3\% &        7.3\% \\
Train on similar        &            0.7\% &            44.5\% &             20.1\% &             6.0\% &              4.3\% &            1.7\% &             0.8\% &        3.1\% \\
60 vs 220M (T5)       &           15.2\% &            85.7\% &             11.4\% &            79.1\% &              3.9\% &           62.5\% &             1.9\% &       13.3\% \\
41 vs. 110M (BERT)   &            4.8\% &            67.0\% &             21.5\% &            41.9\% &              9.2\% &           22.5\% &             4.9\% &        9.0\% \\
110 vs. 340M (BERT)     &            1.1\% &            44.3\% &             36.3\% &            21.9\% &             18.2\% &           11.0\% &             7.3\% &        8.5\% \\
\hline
\end{tabular}
    \caption{All results, with metrics explained in Section \ref{sec:metrics} and Appendix \ref{appendix-sec:comprehensive-results}. Each label is weighted equally.}
    \label{tab:uniformlabel}
\end{table*}

\begin{table*}[]
    \centering
    \begin{tabular}{lrrrrrrrr}
\hline
{} &  $\bar{\Delta}$&  $>$ 1\% &   $<$ -1\% &  $>$ 5\% &  $<$ -5\%  &  $>$ 10\% &  $<$-10\% &  $std(\Delta)$ \\
\hline
Meta-tuned vs QA &            1.2\% &            55.4\% &             35.7\% &            31.2\% &             17.7\% &           15.6\% &            13.6\% &       11.2\% \\
220 vs 770M (T5)      &            6.3\% &            77.4\% &             16.5\% &            51.7\% &              7.0\% &           31.6\% &             4.5\% &        9.0\% \\
Pre-trained vs. Random  & 20.2\% &          89.8\% &              8.5\% &            84.8\% &              6.1\% &           76.6\% &             1.5\% &       15.1\% \\
Ensemble                &            0.1\% &            18.6\% &             20.2\% &             4.3\% &              1.9\% &            1.5\% &             1.2\% &        2.8\% \\
Initialized with QA     &            2.3\% &            59.2\% &             22.5\% &            34.3\% &              9.9\% &           13.9\% &             5.7\% &        7.2\% \\
Train on similar        &            0.6\% &            48.8\% &             25.4\% &             7.3\% &              5.7\% &            1.3\% &             0.9\% &        3.3\% \\
60 vs 220M (T5)       &           12.1\% &            84.6\% &             12.9\% &            73.6\% &              3.5\% &           52.9\% &             2.2\% &       11.6\% \\
41 vs. 110M (BERT)  &            7.0\% &            74.6\% &             13.8\% &            58.5\% &              6.8\% &           31.5\% &             2.9\% &        8.9\% \\
110 vs. 340M (BERT) &            1.1\% &            45.6\% &             36.1\% &            25.5\% &             18.6\% &           10.8\% &             9.3\% &        8.8\% \\
\hline
\end{tabular}
    \caption{All results, with metrics explained in Section \ref{sec:metrics} and Appendix \ref{appendix-sec:comprehensive-results}. Each dataset is weighted equally.}
    \label{tab:uniformdataset}
\end{table*}

\section{Most Relevant Datasets} \label{appendix-sec:most-relevant}

To ensure that we are testing the models' ability to generalize to an unseen tasks, we disallow both training and testing on datasets that are too similar, which is defined as ``having the same set of dataset property tags" (Section \ref{main-sec:data}).
To help interpret how we define unseen tasks, for each dataset that we evaluate on, we try to find the ``most relevant" dataset that the model has seen during the meta-tuning phase, and list it in Table \ref{tab:mostrelevant}.

\begin{table*}[]
    \centering
    \begin{tabularx}{16cm}{|X|X|}
    \hline
        Evaluation Dataset & Most Relevant Training Dataset  \\
    \hline
        SemEval 2016 Task 6, stance classifications on issues like feminism, atheism, etc & SemEval 2019 Task 5, detecting hate speech against women and immigrants \\
    \hline
    SemEval 2019 Task 6, classifying whether the text is offensive & A dataset from Kaggle that classifies sexually explicit comments  \\
    \hline
    SemEval 2019 Task 5, detecting hate speech against women and immigrants & SemEval 2016 Task 6, stance classifications on issues like feminism, atheism, etc \\
    \hline
    TREC, classifying the type the question is asking about (e.g. numbers, acronyms, human/occupations, etc) & AG News, which classifies news into different categories (e.g. sports, world events). \\
    \hline
    SemEval 2019 Task 8, classifying whether the question is asking for subjective opinion, factual information, or simply having a conversation & N/A \\
    \hline
    SUBJ, classifying whether the text contains subjective or objective information & N/A \\
    \hline
    QQP, classifying whether two questions have the same meaning & N/A \\
    \hline 
    \citet{yin-etal-2019-benchmarking} emotion classification, classifying text into 9 emotion types, such as ``joy", ``anger", ``guilt", ``shame", etc. & Classifying whether an IMDB movie review is positive.\\
    \hline
    \citet{yin-etal-2019-benchmarking} situation classification, classifying which disaster situation people are experiencing, e.g. ``regime change", ``crime and violence", and what resource they need, e.g. ``food and water", ``search and rescue". & Classifying (binary) whether a tweet is related to a natural disaster. \\
    \hline
    \citet{yin-etal-2019-benchmarking} topic classification, classifying the domain of an article into domains such as ``family and relationship", ``education", ``business", ``sports" & classifying the domain of a paper abstract into physics, maths, computer sciences, and statistics.\\
    \hline 
    AG News, which classifies news into different categories (e.g. sports, world events). & Abstract Domain classification, classifying the domain of a paper abstract into physics, maths, computer sciences, and statistics. \\
    \hline 
    Abstract Domain classification, classifying the domain of a paper abstract into physics, maths, computer sciences, and statistics. & AG News, which classifies news into different categories (e.g. sports, world events). \\
    \hline
    IMDB movie reviews, classifying whether the user feels positive about the movie & Stock market sentiment, classifying whether a comment is optimistic about the market. \\
    \hline
    CoLA, classifying whether a sentence is grammatical & N/A \\
    \hline
    SemEval 2020 Task 6, classifying whether a sentence contains a definition & N/A \\
    \hline 
    Spam classification, classifying whether a text message is a spam & click-bait classification, classifying whether the title of an article is a clickbait. \\
    \hline 
    SemEval 2018 Task 1, classifying a tweet as one of 4 emotion types \{``sadness", ``joy", ``anger", ``optimism"\} & Classifying whether an IMDB movie review is positive.\\
    \hline
    SemEval 2018 Task 3, classifying whether a tweet is ironic & classifying whether a news title is sarcastic.\\
    \hline
    \end{tabularx}
    \caption{For each dataset that we evaluate on, we list the task in the training split that we consider to be the most relevant. We list ``N/A" if we think that none of the training dataset is particularly relevant. 
    }
    \label{tab:mostrelevant}
\end{table*}

\section{Performance Break Down} \label{appendix-sec:breakdown}
For each model, we average the AUC-ROC scores for each label description for each dataset, and report the results in Table \ref{tab:breakdown}.

\begin{table*}[]
    \centering
    \begin{tabular}{lrrrrr}
\hline
{} &  QA &  QA + Meta &  Meta &  T5 220M &  BERT 340M \\
\hline
Abstract Classification    &      76.9\% &                    84.3\% &        81.2\% &    68.0\% &      85.3\% \\
AG News         &      76.5\% &                    82.0\% &        77.8\% &    69.9\% &      69.5\% \\
Stance (Hillary) &      74.8\% &                    79.8\% &        73.8\% &    69.0\% &      63.2\% \\
Hate Speech            &      59.4\% &                    66.0\% &        64.1\% &    59.6\% &      69.2\% \\
Stance (Feminism)   &      67.8\% &                    71.6\% &        69.1\% &    61.0\% &      64.8\% \\
Stance (Climate)   &      75.8\% &                    81.7\% &        79.6\% &    72.0\% &      76.2\% \\
Emotion Classification$^{*}$  &      67.6\% &                    70.5\% &        68.0\% &    65.0\% &      64.0\% \\
Emotion Classification  (SemEval)       &      81.6\% &                    85.2\% &        81.7\% &    76.1\% &      74.2\% \\
Irony Detection           &      67.9\% &                    83.4\% &        80.2\% &    61.0\% &      64.9\% \\
Stance (Atheism)    &      60.2\% &                    62.4\% &        65.6\% &    55.1\% &      60.9\% \\
QQP                    &      54.1\% &                    61.1\% &        68.6\% &    56.7\% &      66.9\% \\
TREC                                 &      59.3\% &                    63.9\% &        76.4\% &    73.4\% &      66.9\% \\
Stance (Abortion)   &      58.2\% &                    61.3\% &        62.8\% &    60.5\% &      59.5\% \\
Offensive Speech       &      76.6\% &                    80.4\% &        79.5\% &    74.5\% &      80.6\% \\
CoLA                                &      52.3\% &                    49.4\% &        49.8\% &    49.6\% &      50.0\% \\
SUBJ                                &      62.8\% &                    66.8\% &        58.7\% &    54.5\% &      50.2\% \\
Situation Classification$^{*}$  &      73.9\% &                    80.4\% &        79.3\% &    75.5\% &      79.5\% \\
SPAM Detection                       &      57.2\% &                    45.4\% &        35.0\% &    49.3\% &      47.8\% \\
IMDB Movie Review               &      92.9\% &                    94.0\% &        90.5\% &    67.7\% &      84.4\% \\
Topic Classification$^{*}$     &      77.6\% &                    82.7\% &        84.0\% &    77.5\% &      80.7\% \\
Definition Detection                          &      72.8\% &                    73.5\% &        63.9\% &    63.6\% &      60.2\% \\
Question Type Classification  &      75.1\% &                    73.8\% &        59.3\% &    51.8\% &      64.5\% \\
\hline
\end{tabular}
    \caption{Zero shot performance of each model on each dataset. ``QA" means the UnifiedQA model; ``QA + Meta" means meta-tuning with UnifiedQA initialization; ``Meta" means meta-tuning on T5 (770M) parameters. To save space, we use ``*" to denote datasets from \citet{yin-etal-2019-benchmarking}.}
    \label{tab:breakdown}
\end{table*}

\section{Accuracy}

\begin{table*}[]
    \centering
    \begin{tabular}{lrr}
\hline
                              Dataset name &  \#classes &  Accuracy \\
\hline
  2016SemEval6TweetEvalStanceAtheism &            3 &   66 \\
       KaggleNewsTopicClassification &            4 &   64 \\
      2019SemEval6TweetEvalOffensive &            2 &   28 \\
                   2019SemEval8Qtype &            2 &   73 \\
          2018SemEval3TweetEvalIrony &            2 &   39 \\
  2016SemEval6TweetEvalStanceHillary &            3 &   55 \\
                                subj &            2 &   61 \\
                                trec &            6 &   38 \\
                   KaggleQuoraQPairs &            2 &   50 \\
                          definition &            2 &   32 \\
  BenchmarkingZeroshotTopic &           10 &   59 \\
           2019SemEval5TweetEvalHate &            2 &   42 \\
                                cola &            2 &   55 \\
        2018SemEval1TweetEvalEmotion &            4 &   72 \\
 2016SemEval6TweetEvalStanceAbortion &            3 &   64 \\
               KaggleIMDBMovieReview &            2 &   85 \\
  2016SemEval6TweetEvalStanceClimate &            3 &   61 \\
                       KaggleSMSSPAM &            2 &   14 \\
 2016SemEval6TweetEvalStanceFeminist &            3 &   53 \\
\hline
\end{tabular}
    \caption{We report the accuracy of the meta-tuned model for completeness according to the request of the reviewers. 
    However, given that accuracy is very sensitive to thresholding \cite{Zhao2021Calibrate} and is generally unreliable when the labels are imbalanced, these numbers are not likely to be informative. Additionally, to speed up evaluation, we use a subsample of the original test split for some datasets, so these numbers are not directly comparable to those in the other papers either.}
    \label{tab:appendix_accuracy}
\end{table*}

\end{document}